\documentclass[letterpaper, 10 pt, conference]{ieeeconf}  

\IEEEoverridecommandlockouts                              

\usepackage{cite}
\usepackage{amsmath,amssymb,amsfonts}
\usepackage{algorithmic}
\usepackage{graphicx}

\usepackage{textcomp}
\usepackage{xcolor}
\usepackage{url}
\usepackage{booktabs}
\usepackage{tabularx}
\usepackage{subcaption}
\usepackage[hidelinks]{hyperref}
\usepackage{svg}  
\usepackage{cuted}
\usepackage{caption}
\svgpath{{images/}}
\usepackage{float}

\title{\LARGE \bf
Preparation of Papers for IEEE Sponsored Conferences \& Symposia*
}

\author{Dequan Ou$^{1}$, Austin Lopez$^{1}$, Quentin Missinne$^{1}$, Christophe De Wagter$^{1}$ and Guido C. H. E. de Croon$^{1}$
\thanks{*This project was financed by the Dutch Research Council (NWO) VICI personal grant programme under grant number 20663, Air Force Office of Scientific Research (Award Number FA8655-24-1-7061) and European Commission Horizon project SPEAR under grant agreement 101119774.}
\thanks{$^{1}$All authors are with the Micro Air Vehicle Lab of the Faculty of Aerospace Engineering, Delft University of Technology, 2629 HS Delft, The Netherlands
        {\tt\small D.Ou@tudelft.nl}}%
}

\begin{document}
\title{\textbf{Rendering-Free Event Frame Simulation for Fast Reinforcement Learning of Autonomous Robot Navigation}}

\maketitle

\begin{abstract}
Event cameras offer low latency, high dynamic range, and milliwatt-level power consumption, making them ideal for agile robots such as tiny drones navigating through cluttered environments. However, existing simulators generally render photorealistic images before converting them into events, making large-scale reinforcement learning prohibitively slow. We propose an analytic event simulator that exploits what event cameras fundamentally measure---moving contrast edges---to synthesize event streams directly from scene geometry and ego-motion without visual rendering. Our simulator generates a $64{\times}64$ event observation in $17\,\mu$s on a single GPU---over $400\times$ faster than a state-of-the-art rendering pipeline---enabling end-to-end reinforcement learning for goal-directed obstacle avoidance to converge in under 30 minutes---three orders of magnitude faster than training on rendered events. Policies transfer zero-shot to physically based event simulations (ray-traced Isaac Sim). Our policy reaches 83\% success in cluttered courses unseen during training (mid-difficulty traversability, $P_\tau\approx0.57$), at 2.4\,m/s average goal velocity, and retains 70\% in a simulated texture-dense indoor hall. Furthermore, we quantized the policy to a 1.2\,MB int8 network running fully onboard a 56-gram Crazyflie equipped with a Prophesee GenX320 event camera and an OpenMV N6 microcontroller board. Without any fine-tuning, external processing, or positioning, the policy achieves an average success rate of 82\% (5/5, 5/6, 4/6 flights) across three real-world obstacle courses of increasing difficulty. Code: \url{https://github.com/tudelft/rendering-free-event-nav}
\end{abstract}


\section{Introduction}
\begin{figure}[t!]
\centering
\includegraphics[width=0.97\linewidth]{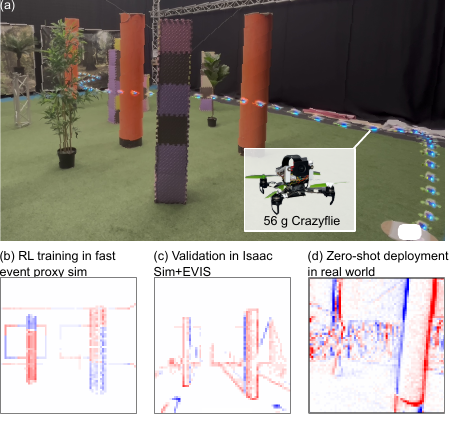}
\caption{(a) A real flight through our test arena: the 56\,g Crazyflie (inset) navigates the clutter fully onboard using a GenX320 event camera and an STM32N6 microcontroller. (b--d) Event frames from the three worlds (blue ON, red OFF): (b) the analytic proxy used for training and (c) the Isaac Sim+EVIS~\cite{shi_evis_2026} rendered validation, both showing a replica of this arena from a matched viewpoint; (d) the real sensor view in flight.}
\label{fig:teaser}
\end{figure}

Flying robots such as tiny drones are attractive for search-and-rescue, inspection, and monitoring in confined, GPS-denied spaces, yet reliable autonomous navigation in such settings remains a fundamental challenge.
Conventional navigation stacks such as Simultaneous Localization and Mapping (SLAM) \cite{cadena_past_2016} demand heavy sensing suites and power-hungry onboard computers that small platforms cannot carry.

Recently, an extremely lightweight approach to navigation, Bee-Nav, has been introduced that achieves long-range navigation with minimal sensing and compute~\cite{ou_efficient_2026}. But for obstacle avoidance, it still relies on depth sensing.
Stereo depth cameras such as the Intel RealSense D435i (72\,g, ${\sim}3.5$\,W) deliver dense, reliable range images and are routinely carried by drones in the 500--1000\,g class---but the camera alone approaches a Crazyflie's entire takeoff weight.
Multi-zone Time-of-Flight arrays, by contrast, are light enough for any nano-drone (sub-gram), yet resolve only $8{\times}8$ zones---too coarse for thin wires, slender poles, or fine vegetative clutter.
On sub-100-gram nano-drones such as the Bitcraze Crazyflie, the Size, Weight, and Power envelope binds tighter still, bounding the entire sensing, processing, and carrier payload to a few dozen grams and roughly a watt.
This rules out not only dedicated depth cameras but also the embedded GPU computers on which conventional event-processing stacks rely (e.g., an NVIDIA Jetson Orin NX module: ${\sim}70$--100\,g, 10--25\,W).
Clutter-resilient obstacle avoidance within this envelope requires rethinking the sensing paradigm.

Event cameras are biologically inspired sensors that have emerged as a compelling modality for autonomous robots~\cite{gallego_event-based_2022}.
Operating asynchronously at microsecond temporal resolution with high dynamic range ($>120$\,dB) and milliwatt-level power consumption, these neuromorphic sensors capture logarithmic brightness changes with negligible motion blur.
These characteristics are particularly advantageous for agile drones, where aggressive rotational maneuvers and unpredictable lighting frequently defeat frame-based cameras.

Event-based obstacle avoidance has accordingly been explored through optical-flow pipelines~\cite{dinaux_faith_2021} and direct end-to-end learning~\cite{bhattacharya_monocular_2024} (Sec.~\ref{sec:related}).
Training robust visuomotor policies with reinforcement learning (RL), however, remains fundamentally bottlenecked by event simulation: standard simulators synthesize events by interpolating pixel intensities across high-frame-rate rendered image sequences~\cite{Rebecq18corl, hu_v2e_2021}, and even recent GPU-accelerated renderers~\cite{shi_evis_2026} throttle the millions of closed-loop environment transitions that end-to-end RL demands.

In this work, we propose an approach to sidestepping rendering entirely.
Observing that event cameras fundamentally respond to moving contrast edges, we propose an analytic event proxy that synthesizes event streams directly from scene geometry and ego-motion.
Event generation costs $17\,\mu$s per frame, and end-to-end RL for goal-directed obstacle avoidance converges in under 30 minutes.
The learned policy transfers zero-shot to a physically based event simulation (83\% success in unseen cluttered rooms). Quantized to a 1.2\,MB int8 network running fully onboard a 56-gram Crazyflie with a Prophesee GenX320 event camera, it further achieves 100\%, 83\%, and 67\% success (5/5, 5/6, 4/6 flights) in three distinct, increasingly difficult real environments, with all sensing, processing, and state estimation onboard.

\vspace{1mm}
\noindent The main contributions of this work are:
\begin{itemize}
    \item an ultra-fast, rendering-free analytic event proxy for navigation tasks that trains avoidance policies three orders of magnitude faster than rendered-event pipelines;
    \item a sim-to-sim-to-real transfer study showing that policies trained purely in the proxy transfer zero-shot to photorealistic Isaac Sim and to a real Prophesee GenX320;
    \item the first fully onboard, event-driven goal-directed navigation on a 56-gram Crazyflie, with all perception and inference on a sub-watt microcontroller.
\end{itemize}

\section{Related Work}
\label{sec:related}



\subsection{Event-based drone flight}

Prior work on event-based flight spans indirect pipelines---which first reconstruct an intermediate quantity such as optical flow, time-to-contact, or depth, and act on it---and direct methods that map event streams to commands.
On the indirect side, Falanga~et~al.~\cite{falanga_dynamic_2020} demonstrated dodging of fast dynamic obstacles with a hand-designed ego-motion-compensated detection pipeline (learned variants followed~\cite{sanket2020evdodgenet}), Dinaux~et~al.~\cite{dinaux_faith_2021} pioneered onboard \emph{static}-obstacle avoidance by estimating the focus of expansion from event-based optic flow (80\% on a frontal pole), and Hagenaars~et~al.~\cite{hagenaars_-device_2025} learn monocular event-based depth self-supervised, on board a small drone, for obstacle avoidance.
On the direct side, Bhattacharya~et~al.~\cite{bhattacharya_monocular_2024} presented a static-obstacle avoidance for quadrotors from a monocular event camera, pre-training an events-to-control policy on approximated simulated events and fine-tuning its perception component with real events-and-depth data. Messikommer~et~al.~\cite{messikommer_approximate_2026} learned commanded-direction avoidance in clutter by imitating a privileged teacher while minimizing event rendering, deploying on an NVIDIA Jetson.

These systems convincingly demonstrate the latency and dynamic-range advantages of events, but they execute their networks off-board or on onboard computers drawing typically 10\,W or more (NVIDIA Jetson family), an order of magnitude beyond the power and payload budget of a nano-drone.
At the extreme low-power end, Paredes-Vall\'es~et~al.~\cite{paredes-valles_fully_2024} achieved the first fully neuromorphic vision-to-control loop on a flying drone, running spiking networks on the Intel Loihi research chip for optical-flow-based flight, landing, and hovering.
Goal-directed navigation through cluttered environments, computed fully on board a commercially available microcontroller, has remained open---this is the capability we demonstrate.

\begin{figure*}[t]
\centering
\includegraphics[width=0.98\textwidth]{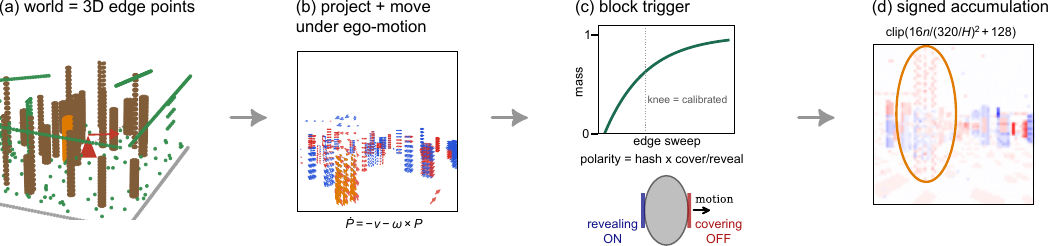}
\caption{The rendering-free event-proxy pipeline: one obstacle (orange) is traced through every stage. (a) A scene is only 3D edge points on depth discontinuities (trunks brown, walls gray, props green; the red triangle is the camera, the arrow its velocity). (b) Visible points project to the image plane and move under ego-motion; arrow color is polarity. (c) The two firing mechanisms: per-cell event mass follows a saturating trigger (the block-level aggregate of binary pixel crossings; its knee---the displacement at which a cell saturates---is fitted to real sensor recordings), and polarity is a hashed contrast class times the covering/revealing side of the moving edge. (d) Signed accumulation yields the $64{\times}64$ DIFF3D frame the policy observes; the traced obstacle's streak is circled.}
\label{fig:pipeline}
\end{figure*}

\subsection{Event camera simulation}
Event datasets exist for driving and agile flight~\cite{zhu_multi_2018, gehrig_dsec_2021, zhong_event-based_2026}, but fixed recordings cannot provide the interactive experience that closed-loop reinforcement learning requires; simulation therefore plays a central role, and existing event simulators are generally render-and-convert pipelines.
One family converts conventional video, optionally with frame interpolation: v2e~\cite{hu_v2e_2021}, Vid2E~\cite{gehrig_video_2020}, and V2CE~\cite{zhang_v2ce_2024} produce realistic events but replay fixed footage and cannot react to an agent's actions.
A second family couples an event model to a rendering engine: ESIM~\cite{Rebecq18corl} drives an adaptive log-intensity interpolator from a renderer, a DVS plugin brings events to Gazebo~\cite{kaiser_towards_2016}, and circuit-level models such as IEBCS~\cite{joubert_iebcs_2021} add characterized sensor noise to rendered frames.
Most recently, EVIS~\cite{shi_evis_2026} accelerates the render stage inside Isaac Sim with motion-vector warping, approaching real-time throughput; we adopt it as the physically based reference simulation for our sim-to-sim transfer evaluation.
All of these must produce images before events.

Render-free event generation remains largely unexplored.
GERD~\cite{pedersen2024gerd} synthesizes events from controlled 2D affine transforms of prototypical objects for equivariance studies, without 3D scenes or a closed loop.
Messikommer~et~al.~\cite{messikommer_approximate_2026} attacked the same bottleneck by routing around event simulation rather than cheapening it: event rendering is confined to a one-time offline dataset used to learn a task representation, and during policy search a network approximates that representation directly from lightweight privileged state, so no events are rendered at all (a $28\times$ speedup).
In contrast, we make closed-loop event generation itself analytic---no images or offline rendered dataset anywhere---and train end-to-end with RL on event inputs rather than imitating a privileged expert; the approximation is validated by transfer to a physically based event simulation and to a real sensor, zero-shot and without fine-tuning.

\subsection{Autonomous nano-drones and learned flight}
\emph{Autonomous nano-drones.}
Full autonomy under a gram-and-milliwatt budget has so far been reached with minimal, carefully engineered pipelines.
The DelFly Explorer~\cite{de_wagter_autonomous_2014} achieved autonomous obstacle avoidance on a 20\,g flapping-wing MAV, which included a 4\,g onboard stereo system.
Sniffy Bug~\cite{duisterhof_sniffy_2021} achieved fully autonomous gas-source seeking with a swarm of 37.5-gram nano-quadcopters in cluttered, GPS-denied spaces.
On the perception side, the PULP line squeezed deep networks onto milliwatt processors: PULP-DroNet~\cite{palossi_64-mw_2019} demonstrated closed-loop DNN-based visual navigation at 64\,mW, follow-up work brought sim-to-real nano-drone racing~\cite{lamberti_sim--real_2024}, fully onboard human-pose estimation~\cite{palossi_fully_2022}, and NanoFlowNet~\cite{bouwmeester_nanoflownet_2022} computes dense optical flow in real time for obstacle avoidance.
All of these rely on frame-based or range sensing and on supervised perception modules or hand-designed controllers tuned in simulation---none learns control end-to-end, and none uses events.

\emph{Learned vision-based flight.}
On standard-size quadrotors, simulation-trained policies have achieved remarkable results: Loquercio~et~al.~\cite{loquercio_learning_2021} learn high-speed flight in the wild from depth images by imitating a privileged expert; Song~et~al.~\cite{song_learning_2023} add perception awareness through learning-by-cheating; Kulkarni~et~al.~\cite{kulkarni_semantically-enhanced_2023} and MAVRL~\cite{yu_mavrl_2024} learn depth-latent collision prediction and complexity-adaptive speed.
Closer to our end-to-end setting, SkyDreamer~\cite{verraest_skydreamer_2025} maps pixel-level representations directly to motor commands with model-based RL, achieving onboard sim-to-real drone racing; the direct event-based methods discussed above~\cite{bhattacharya_monocular_2024, messikommer_approximate_2026} bring this paradigm to event cameras.
Across all of these, training uses rendered or real visual data and deployment assumes desktop-class or Jetson-class compute.
To our knowledge, no prior system combines end-to-end reinforcement learning, event-based vision, and fully onboard microcontroller inference at nano-drone scale; enabling exactly this combination---by removing the event-simulation bottleneck that separates these two bodies of work---is the aim of this paper.

\section{Methodology}
Our pipeline has three stages, summarized in Fig.~\ref{fig:teaser}: an analytic event proxy that trains RL policies without rendering (Sec.~\ref{sec: EPG} and Sec.~\ref{sec:RL}), zero-shot validation in a physically based event simulation (Sec.~\ref{sec:isaacval}), and quantized deployment on a microcontroller for real-world flight experiments (Sec.~\ref{sec:deploy}).

\subsection{Event proxy generation}
\label{sec: EPG}
\textbf{Sensor model.}
The target sensor is a GenX320 in DIFF3D mode: each native pixel fires when its log-intensity change crosses a contrast threshold, and the firmware accumulates a \emph{signed event count} per pixel over successive 20\,ms integration windows (50\,Hz), block-averaged to the $H{\times}H$ observation.
The frame the policy sees therefore contains no intensities and no timestamps---only where events fell within the window, their net polarity, and their local density; opposite-polarity events cancel (a pixel receiving one ON and one OFF event reads identically to one receiving none).
This is the entire process the proxy attempts to reproduce.

\textbf{Scene and event model} (Fig.~\ref{fig:pipeline}a--d).
In the proxy simulator, a scene is not represented by meshes and materials but by a set of 3D edge points sampled on depth discontinuities such as trunk silhouettes, wall outlines, and prop contours---the structures a moving event camera typically reports---held in fixed-size point buffers for batched evaluation across 512 parallel environments.
Surface \emph{texture} (a printed wall firing events across its face) is not modeled on walls, though the world-anchored hashed texture used for the ground extends to any surface by adding points.
Given camera velocity and rotation rates $(\mathbf{v}, \boldsymbol{\omega})$, a world-static point at camera-frame position $\mathbf{P}$ moves as $\dot{\mathbf{P}} = -\mathbf{v} - \boldsymbol{\omega}\times\mathbf{P}$ and sweeps the image-plane segment
\begin{equation}
\boldsymbol{\delta} \;=\; \pi\!\left(\mathbf{P} + \dot{\mathbf{P}}\,\Delta t\right) - \pi(\mathbf{P}),
\qquad \Delta t = 20\,\mathrm{ms},
\label{eq:sweep}
\end{equation}
where $\pi$ is the pinhole projection.
The linear sweep is exact under translation and a small-angle approximation under rotation; over a 20\,ms window at $\dot\psi_{\max}=1$\,rad/s the arc--chord deviation is below $10^{-2}$\,px (level flight excludes roll).
Events deposit along this streak---the spatial footprint a moving edge leaves in a count frame---with per-point mass and per-cell output
\begin{equation}
m = g\,\min\!\left(\|\boldsymbol{\delta}\|/\kappa,\; 1\right),
\qquad
I = \mathrm{clip}\!\left(\frac{16\,n}{(320/H)^2} + 128\right),
\label{eq:mass}
\end{equation}
where $g$ (gain) is the event count a fully-firing edge point deposits---playing the role of typical edge contrast, the per-edge quantity geometry cannot supply---$\kappa$ (knee) the image displacement at which an output cell's native block is fully swept, and $n$ the signed sum of deposited mass in a cell (one cell of the $64^2$ observation = the block mean of $5{\times}5$ native pixels); the clip is to the sensor's 8-bit range, matching the firmware chain of native-pixel events followed by block-mean downsampling. Both $g$ and $\kappa$ are fitted by the calibration check below.
The saturating ramp in Eq.~\eqref{eq:mass} is the block-level aggregate of binary threshold crossings: a single native pixel either fires or does not, so an output cell covering a $5{\times}5$ native block responds in proportion to how many were swept, saturating at full coverage---faster motion lengthens the streak rather than brightening it.

\textbf{Polarity} (Fig.~\ref{fig:pipeline}c) is the product of two signs.
A \emph{contrast class} is hashed from each point's world position at 0.8\,m granularity: true albedo contrast is unknowable from geometry alone, but it is spatially coherent (an edge looks like one edge) and temporally fixed (it never flips between frames).
A \emph{covering/revealing} factor supplies the physics: using per-point outward normals, an edge advancing along its projected normal covers background and fires one polarity; receding reveals it and fires the other.
Each moving edge deposits a \emph{single-polarity} streak (a step edge sweeps every pixel in the same brightness direction; the opposite sign comes from the object's other edge), so forward flight produces coherent polarity around an expanding silhouette while yaw splits leading and trailing edges---matching real recordings.

Three secondary effects complete the frame: a foreshortening weight $\sin\angle(\mathbf{n},\mathbf{r})$, between a point's outward normal $\mathbf{n}$ and its view ray $\mathbf{r}$, silences edges whose normal points along the view ray---such points sweep pixels like any other, but a surface seen face-on presents no visible contrast step, a coarse z-buffer over the point set provides occlusion, and the ground---a continuous textured surface rather than an object---is shaded analytically per pixel below the horizon, a world-anchored hash deciding which texture cells fire under the local optic flow.
Clutter props are contour sketches: screens as border points plus a content blob, trusses as hollow rail-and-rung lattices, plants as a stem plus a canopy cloud whose vertical spread crosses polarity cells, reproducing the mixed-sign foliage of real footage.
Per-episode \emph{render-statistics domain randomization} varies gain, firing knee, polarity grain, ground-texture fill, and dilation.

\textbf{Calibration.}
The proxy's five physical knobs are pinned by a calibration check: five summary statistics (background fraction, 50th/90th-percentile magnitudes, blob size, ON-polarity fraction) computed identically on real GenX320 recordings and proxy output under matched motion, each knob dominating one statistic.
The check is re-run after every proxy change; configurations that drift outside tolerance are rejected.
Calibration is sensor identification, not policy adaptation: it fits five scalars to a single 10-second recording of the sensor moved by hand in an office---not in any flight environment---and no flight or task data from any target domain is used in training; the policy remains zero-shot.

\subsection{Reinforcement learning}
\label{sec:RL}
\textbf{Task.}
Goal-directed navigation in clutter: from a random spawn (position, heading, and initial speed all randomized, with a $\sim$1\,m clearance guarantee) in a $22{\times}22$\,m room with randomized obstacle density and props, fly to a goal on the far side of the room.
An episode ends on reaching the goal ($<1$\,m), collision, leaving the room, or timeout.

\textbf{Observations and actions.}
The policy receives the $64{\times}64$ single-channel DIFF3D frame (three stacked timesteps) and a 4-vector $[\cos\beta, \sin\beta, \dot\psi/\dot\psi_{\max}, \|\mathbf{v}\|/v_{\max}]$, where $\beta$ is the body-frame goal bearing; speed and yaw rate come from onboard state estimation.
A $\tanh$ head outputs yaw rate ($\dot\psi_{\max}{=}1$\,rad/s through a $\tau{=}0.08$\,s lag) and forward speed ($v \in [0, v_{\max}]$, slew-limited at 3\,m/s$^2$).

\textbf{Decision timing.}
The sensor integrates successive 20\,ms windows continuously; each decision takes the three most recent windows as input, so the stride between stacked frames equals the decision period.
The decision rate is set by the N6's inference and frame-assembly latency---not by the sensor---and because the stride ties the observation to that rate, training must use the deployed rate: 13.3\,Hz, realized as alternating 7/8-tick holds on the 100\,Hz simulation.

\textbf{Reward.}
The reward, evaluated per 100\,Hz simulation tick, is
\begin{equation}
\begin{aligned}
r ={}& 5\,\Delta d \;+\; \Delta\Phi_c \;+\; \Delta\Phi_a \\
&- 0.05\,\mathrm{relu}(|\beta| - \tfrac{\pi}{2})\,\mathbb{1}[c{>}3] - 10\,\mathbb{1}_{\mathrm{crash}} + 30\,\mathbb{1}_{\mathrm{goal}},
\end{aligned}
\end{equation}
with $\Delta d$ the per-tick decrease in goal distance, $c$ the distance to the nearest obstacle surface, and $\beta$ the heading error to the goal bearing.
Both shaping terms are \emph{potential-based}, applied as differences of $\Phi_c = -2\,\mathrm{relu}(0.8 - c)^2$ (clearance) and $\Phi_a = -0.5\,(1-\cos\beta)\,\mathbb{1}[c{>}3]$ (alignment), so they telescope and cannot be farmed; the gates confine heading pressure to open space, away from avoidance.

\textbf{Training.}
Standard clipped-surrogate PPO (Table~\ref{tab:ppo}) with a compact convolutional trunk (16/32/64 channels; 1.12\,M parameters).
\begin{table}[t]
\caption{Training configuration (identical for all policies).}
\label{tab:ppo}
\centering\footnotesize\setlength{\tabcolsep}{4pt}
\begin{tabular}{llll}
\toprule
parallel envs & 512 & rollout length & 128 \\
epochs / minibatches & 2 / 8 & discount $\gamma$ & 0.99 \\
GAE $\lambda$ & 0.95 & clip $\epsilon$ & 0.2 \\
learning rate (Adam) & $3{\times}10^{-4}$ & entropy coef.\ & $5{\times}10^{-3}$ \\
value coef.\ & 0.5 & reward scale & 0.05 \\
total steps & 40\,M & wall-clock & 34\,min \\
\bottomrule
\end{tabular}
\end{table}
The vectorized proxy sustains 19--21k environment steps/s on one RTX~4090; a 40M-step run completes in 34 minutes, with training success crossing 80\% within the first 8 minutes.

\begin{figure}[t]
\centering
\includegraphics[width=0.9\linewidth]{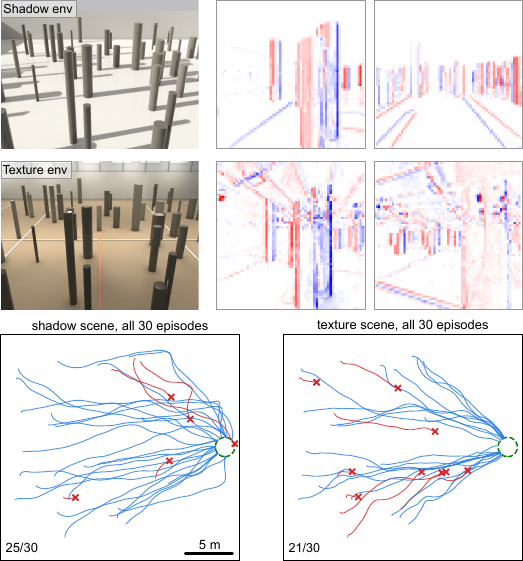}
\caption{Simulation evaluation under the shared action interface. Top and middle rows: the two appearance conditions with identical obstacle layouts---the \emph{shadow} scene (bare surfaces, directional light) and the \emph{texture} scene (court markings, panels, window bands under diffuse light)---as ray-traced views (left) and simulated $64{\times}64$ DIFF3D observations (blue ON, red OFF); the texture scene adds background events the shadow scene lacks. Bottom: all 30 evaluation episodes of our policy in each condition (every episode a distinct procedurally generated layout; blue = reached, red $\times$ = failure, dashed circle = goal), 25/30 and 21/30.}
\label{fig:baselines}
\end{figure}

\subsection{Sim-to-sim validation}
\label{sec:isaacval}
Before real flight, policies are evaluated zero-shot in NVIDIA Isaac Sim with the EVIS event-camera plugin~\cite{shi_evis_2026}: ray-traced HDR keyframes with motion-vector warping to 240--400\,Hz and a log-intensity threshold event model with sensor noise, at the native $320{\times}320$ resolution.
The raw event stream is then converted to the policy's observation exactly as the hardware firmware does: signed per-native-pixel counts over free-running 20\,ms windows, the per-pixel saturating map of Eq.~\eqref{eq:mass} applied \emph{at native resolution}, then the firmware's block-mean downsample to $64^2$---so proxy, rendered, and real frames share one observation format.
The single scale in this map is fitted once so that the rendered frames' active-pixel p90 magnitude matches a real GenX320 recording (EVIS's absolute event rates differ from the physical chip); like the proxy's calibration check, this is sensor identification for the rendered sensor, fitted to a recording rather than to task performance.
(The proxy itself renders directly on the $64{\times}64$ grid; native resolution enters only through the calibrated constants of Eq.~\eqref{eq:mass}.)
Generating events at native resolution before downsampling is essential: event firing is threshold-nonlinear, so a pole one native pixel wide fires correctly at $320^2$ but vanishes from a $64^2$ render.
None of the Isaac scenes---tree fields, textured walls, solid-geometry props---appear in proxy training; this stage functions as a rendered-world holdout.

Beyond visual inspection (Fig.~\ref{fig:teaser}), Table~\ref{tab:framestats} compares frame statistics across the policy's operational worlds.
The statistics separate into two families: per-pixel \emph{shape} statistics---polarity balance, magnitudes, adjacency---which the proxy matches closely, and \emph{density}, where every simulator sits below reality.
Lower density is a modeled economy, not sensor unrealism: the proxy fires only on task-relevant structure, omitting ambient texture, and even the physically based renderer reaches under half of real density in the shadow scene; the texture condition, whose statistics track the real recording, prices this omission (Sec.~\ref{sec:exp-ood}).

\begin{table}[t]
\caption{Event-frame statistics of the policy's operational worlds: proxy training scenes (champion flights, domain randomization active), the two Isaac evaluation conditions, and a real in-flight GenX320 recording. Net-count deviations from the 128 background; block variance over $4{\times}4$-pixel sums; trajectories differ across rows.}
\label{tab:framestats}
\centering\footnotesize\setlength{\tabcolsep}{2.5pt}
\begin{tabular}{lccccc}
\toprule
 & active px & ON share & p50/p90 & blob & blk.\ var. \\
\midrule
proxy (training) & 15\% & 37\% & 2 / 6 & 1.19 & 216 \\
Isaac+EVIS, shadow & 25\% & 40\% & 5 / 14 & 1.59 & 1801 \\
Isaac+EVIS, texture & 54\% & 40\% & 3 / 13 & 1.61 & 2882 \\
real GenX320 & 66\% & 39\% & 3 / 14 & 1.54 & 6947 \\
\bottomrule
\end{tabular}
\end{table}

\subsection{Real-world deployment}
\label{sec:deploy}
The policy that flies is the trained policy---no finetuning, distillation, or architecture change: it is exported through ONNX, quantized to int8 post-training on closed-loop task observations (training itself stays fp32), and compiled for the STM32N6's Neural-ART NPU (convolutions and dense layers on the NPU; frame assembly, normalization, and the output $\tanh$ on the Cortex-M55), yielding a 1.2\,MB binary.
Parity is verified at every step: the Torch port is numerically equivalent to the JAX network ($<5{\times}10^{-6}$), the int8 policy's closed-loop success matches fp32 within evaluation noise in both the proxy and Isaac, and replaying in-flight logged frames through the deployed binary reproduces the reference actions (correlation 1.00)---a pre-flight regression check.

\begin{figure}[t]
\centering
\includegraphics[width=0.9\linewidth]{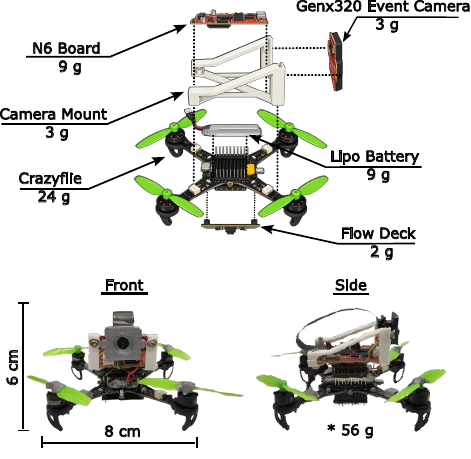}
\caption{The 56\,g flight platform. Top: exploded view---Crazyflie 2.1 base (24\,g), LiPo battery (9\,g), OpenMV N6 board (9\,g), Prophesee GenX320 event camera (3\,g), printed camera mount (3\,g), and flow deck (2\,g) for velocity estimation. Bottom: assembled, front and side. The 56\,g total includes these components, cables, and a boost converter.}
\label{fig:drone}
\end{figure}

The N6 companion board and the Crazyflie exchange messages over a two-wire
UART link (115\,200\,baud rate); the N6 runs only the policy, while the
Crazyflie's STM32F405 runs the low-level stabilization
(Fig.~\ref{fig:drone}). Framing carries a sync header and checksum, so corrupted packets are discarded rather than acted on.

\textbf{Downlink (N6$\,\to\,$Crazyflie).}
At every decision step the policy emits a two-dimensional action
$\mathbf{a}=(a_0,a_1)\in[-1,1]^2$. The Crazyflie
firmware maps them to high-level setpoints: $a_0$ to a yaw-rate command
$\dot\psi = a_0\,\dot\psi_{\max}$ and $a_1$ to a forward speed
$v = \tfrac12(a_1{+}1)\,v_{\max}$, tracked at a fixed cruise height by the
onboard velocity and attitude controllers. The network thus
outputs normalized, hardware-agnostic commands; all physical scaling and
stabilization live on the Crazyflie.

\textbf{Uplink (Crazyflie$\,\to\,$N6).}
The only non-visual input to the policy is a four-dimensional context vector
supplied by the Crazyflie:
$[\,\cos\beta,\ \sin\beta,\ \dot\psi/\dot\psi_{\max},\ |v|/v_{\max}\,]$.
Here $\beta$ is the \emph{egocentric bearing to the goal}, computed on the
Crazyflie entirely from its onboard state estimate---flow-deck odometry against
the goal position fixed at takeoff; no external positioning is used---and
$\dot\psi,\,|v|$ are the current yaw rate and horizontal speed from the same
estimator. The event image therefore carries all obstacle information, while
the vector carries only \emph{where the goal is} and \emph{how the drone is
currently moving}---no map, waypoints, or obstacle state are provided.
The two loops run asynchronously: the policy executes at the measured
13.3\,Hz on the N6, consuming the most recent context vector,
while the Crazyflie holds the last received action and closes its inner loop at
500\,Hz. This zero-order hold, together with the actuator lag,
low-passes the policy's step-to-step commands.

\section{Experiments}

\begin{table}[t]
\caption{Measured event-generation and training throughput (RTX 4090, identical scenes).}
\label{tab:speed}
\centering
\setlength{\tabcolsep}{2.5pt}\footnotesize
\begin{tabular}{lccc}
\toprule
 & sensor sim & closed loop & wall-clock to \\
 & ($64{\times}64$ frame) & (policy steps/s) & trained policy \\
\midrule
Rendered (Isaac+EVIS) & 6.9\,ms & $\sim$2.5 & $\sim$200 days$^{\dagger}$ \\
Analytic proxy (ours) & 17\,$\mu$s$^{\ddagger}$ & 19--21k & \textbf{34\,min} \\
\bottomrule
\multicolumn{4}{l}{\footnotesize $^{\dagger}$extrapolated from the measured GPU-day (195k steps).}\\
\multicolumn{4}{l}{\footnotesize $^{\ddagger}$full observation path (bare event splat 4\,$\mu$s).}
\end{tabular}
\end{table}

\subsection{Evaluation protocol and metrics}
\label{sec:exp-baselines}
We compare four methods under the identical Isaac protocol, summarized in Table~\ref{tab:baselines}: $n{=}30$ as three fixed-seed batches of ten episodes---every episode a distinct procedurally generated layout, with layouts, spawns, and goals identical across methods (scenes contain tree poles plus lab props---screens, trusses, plants; $P_\tau\approx0.57$)---and every method flying the same action interface ($v\in[0,3.5]$\,m/s at 13.3\,Hz).
Each layout is evaluated in two appearance conditions (Fig.~\ref{fig:baselines}): a \emph{shadow} scene (bare surfaces under one directional light, so floor events arise mainly from cast-shadow boundaries) and a \emph{texture} scene (court markings, wall panels, and window bands under diffuse, shadow-free light).
Unlike comparisons that juxtapose success rates as reported in different papers under different setups, every method here is evaluated on the same episodes, seeds, dynamics, and metrics, so differences are attributable to the sensing modality and training method alone.
\label{sec:exp-avoidbench}
No benchmark currently accepts event-camera input, so to express our results in recognized metrics we adopt the metrics of AvoidBench~\cite{yu_avoidbench_2023}, computed identically in our Isaac protocol: success rate (SR), average goal velocity ($AGV = d_{\min}/t_{\mathrm{trial}}$, with $d_{\min}$ the A*-shortest free path), path optimality ($PO = (d_{\mathrm{trav}}-d_{\min})/d_{\min}$), mission progress (MP), and environment traversability (mean collision-checker free-flight distance from random poses, normalized by drone diameter).

\subsection{Zero-shot transfer}
\label{sec:exp-ood}
Our policy---one checkpoint, used unchanged for every result in this paper, including the real flights---carries a single unbroken ladder from training to hardware: 83\% in rendered Isaac scenes it has never seen (Table~\ref{tab:baselines}), 70\% in the texture scene, and 100/83/67\% in three real arenas (Sec.~\ref{sec:exp-real})---each stage evaluated zero-shot, with no fine-tuning, adaptation, or per-domain data anywhere in the chain.
The per-condition ordering in simulation predicted the ordering observed in the real world.
In AvoidBench terms, at normalized traversability $P_\tau \approx 0.57$ (mid-difficulty range) our policy achieves SR 83\%, AGV 2.38\,m/s, PO 3.6\%, and MP 87.9\% in the shadow scene, and SR 70\%, AGV 1.62\,m/s, MP 76.3\% in the texture scene; for reference, the depth-based planners evaluated in AvoidBench report AGV between roughly 1 and 3\,m/s with comparable path optimality on desktop-class compute, while our numbers are produced from events alone by a 1.2\,MB int8-deployable network.
We stress this is a metrics-level, not benchmark-level, comparison: the environments differ, and the published baselines consume depth, not events.

The texture scene dresses the same rooms as a sports hall (court floor markings, panelled walls with window bands, ceiling trusses and lamps), producing event streams far denser than anything the training proxy generates.
The result is a clean dissociation (Table~\ref{tab:baselines}): the appearance-invariant depth policy is essentially unchanged (100\%$\to$97\%), while our policy degrades gracefully (83\%$\to$70\%), preserving its margin over every event-based alternative.
The gap is thus attributable to event-\emph{texture} distribution shift, not geometry; the shift is visible in the frames themselves, active pixels doubling from shadow to texture (Table~\ref{tab:framestats}).
The deployed int8 policy tracks its fp32 source here as well (19/30 vs.\ 21/30): consistently a touch lower, but not a meaningful contributor to the texture gap.

\subsection{Baseline comparisons}
We compare to the following baselines.
\textbf{(i) Privileged depth}: the same PPO trained on analytic ground-truth depth images (97\% training success in the same budget) transfers at 100\%/97\%---a true upper bound, as expected for noiseless privileged geometry, and essentially unchanged across the appearance shift (one episode).
Our monocular event policy reaches 83\%/70\% against that ceiling.
\textbf{(ii) Hand-designed reactive control}: an insect-inspired flow-balancing controller~\cite{serres_optic_2017, milde_bioinspired_2015} with goal-bearing bias, using the same event frames, achieves 13\% (shadow) and 17\% (texture) -- reactive strategies survive only the sparsest layouts and collapse in dense clutter, while the learned policy handles both.
\textbf{(iii) Rendered-events training, one GPU-day}: PPO trained \emph{in the rendering loop} on closed-loop Isaac+EVIS events for a full 24-hour GPU-day---$42\times$ our entire training wall-clock, with the further advantage of training in the evaluation domain itself---collects 195k environment transitions (versus our 40M in 34 minutes) and reaches 40\% (shadow) / 53\% (texture) on the shared evaluation episodes. Per environment transition this learner is remarkably \emph{sample}-efficient (40\% success from 0.5\% of our training steps); its ceiling is purely the renderer's throughput, which is precisely the bottleneck the analytic proxy removes.
Table~\ref{tab:speed} quantifies the underlying economics: the analytic proxy generates a $64{\times}64$ event observation in 17\,$\mu$s where the rendered pipeline needs 6.9\,ms---over $400\times$---which compounds through the closed loop (19--21k vs.\ $\sim$2.5 policy steps/s) into 34 minutes versus an extrapolated ${\sim}200$ GPU-days for the same 40M-step curriculum.
Profiling the proxy training loop attributes 33\% of wall-clock to event generation, 44\% to flight dynamics, 21\% to learning updates, and 1\% to the policy forward pass---event synthesis is not even the largest component.
The rendered pipeline inverts this profile: ${\sim}96\%$ of its per-step cost is producing the event frame, ${\sim}3\%$ learning.

\begin{table}[t]
\caption{Baseline comparison in the Isaac protocol ($n{=}30$, identical seeds, props). AvoidBench metrics: success rate, average goal velocity (m/s, shadow scene), path optimality (excess over the A*-shortest path; successes only).}
\label{tab:baselines}
\centering
\setlength{\tabcolsep}{2.5pt}\footnotesize
\begin{tabular}{lcccc}
\toprule
 & input & SR (shadow) & SR (texture) & AGV \\
\midrule
PPO, privileged depth & depth & 100\% & 97\% & 2.19 \\
\midrule
\textbf{PPO, analytic proxy (ours)}$^{a}$ & events & \textbf{83\%} & \textbf{70\%} & 2.38 \\
Flow balancing (reactive) & events & 13\% & 17\% & 1.10 \\
PPO, rendered, 1 GPU-day & events & 40\% & 53\% & 1.78 \\
\bottomrule
\multicolumn{5}{l}{\footnotesize $^{a}$All methods share the deployed action interface}\\
\multicolumn{5}{l}{\footnotesize ($v{\in}[0,3.5]$\,m/s at 13.3\,Hz); scenes: Sec.~\ref{sec:exp-baselines}.}
\end{tabular}
\end{table}


\subsection{Real-world flight results}
\label{sec:exp-real}
\begin{figure}[t]
\centering
\includegraphics[width=0.94\linewidth]{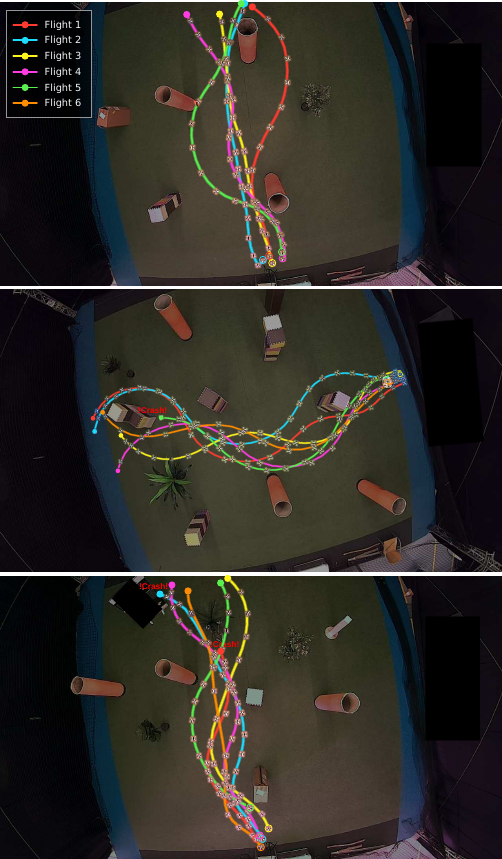}
\caption{All real flights, overhead composites. Top: environment 1 (poles and boxes), 5/5 reached. Middle: environment 2 (extended course with block towers, plants, and screens), 5/6. Bottom: environment 3 (compact arena with tight passages), 4/6; crashes marked. All flights fully onboard---GenX320 events, int8 policy on the N6, flow-based velocity---with no external processing or positioning.}
\label{fig:realtraj}
\end{figure}
The trained policy is evaluated on obstacle courses in the lab: poles, box- and cylinder-shaped obstacles, plants, banners, and plants raised on a beam structure populate the flight arena, with the drone placed on one side and the goal 5--10\,m away on the other. Each flight takes off, crosses the course fully autonomously, and counts as a success only if the drone reaches the goal without touching any obstacle.
Across three distinct arenas our policy---the same network evaluated throughout this section, deployed as a 1.2\,MB int8 binary at 13.3\,Hz---reaches its goal in \textbf{5/5, 5/6, and 4/6} flights (100\%, 83\%, 67\%; Fig.~\ref{fig:realtraj}), with every computation onboard.
Pole-dominated scenes transfer essentially losslessly, while arenas containing organic and structural props (plants, screens, block towers) cost success at a rate consistent with the props-condition Isaac results.
Both crashes in the densest arena occur within cluttered pockets rather than in open space, consistent with the failure signature observed in dense-room simulation episodes. 

\section{Conclusion}
We presented a rendering-free route to event-based drone navigation: an analytic simulator that synthesizes calibrated event observations from scene geometry and ego-motion at $17\,\mu$s per frame, training goal-directed obstacle avoidance with model-free RL in 34 minutes on one GPU.
A single policy trained entirely in this proxy transfers zero-shot to ray-traced event simulation (83\% in unseen clutter; 70\% under a visual-realism shift) and, quantized to a 1.2\,MB int8 network, flies a 56\,g Crazyflie fully onboard at 82\% average success across three real arenas---no external sensing or processing beyond the goal coordinate fixed at takeoff.
Under an identical action interface it approaches a privileged ground-truth-depth upper bound while cutting the training cost of rendered-event RL by orders of magnitude.
The result argues a general point: for sensors whose simulation cost lies in photometry, an observation model calibrated to the sensor's output statistics can replace rendering.
Full attitude, faster control rates, and dynamic scenes are natural next steps.

\section{Acknowledgement}
The authors would like to thank Chaoxiang Ye, Yilun Wu, Hang Yu, Shenqi Wang, Rafael P\'{e}rez and Elijah Ang for their insights during this project.

\bibliographystyle{IEEEtran}
\bibliography{references}

\begin{thebibliography}{10}
\providecommand{\url}[1]{#1}
\csname url@samestyle\endcsname
\providecommand{\newblock}{\relax}
\providecommand{\bibinfo}[2]{#2}
\providecommand{\BIBentrySTDinterwordspacing}{\spaceskip=0pt\relax}
\providecommand{\BIBentryALTinterwordstretchfactor}{4}
\providecommand{\BIBentryALTinterwordspacing}{\spaceskip=\fontdimen2\font plus
\BIBentryALTinterwordstretchfactor\fontdimen3\font minus
  \fontdimen4\font\relax}
\providecommand{\BIBforeignlanguage}[2]{{%
\expandafter\ifx\csname l@#1\endcsname\relax
\typeout{** WARNING: IEEEtran.bst: No hyphenation pattern has been}%
\typeout{** loaded for the language `#1'. Using the pattern for}%
\typeout{** the default language instead.}%
\else
\language=\csname l@#1\endcsname
\fi
#2}}
\providecommand{\BIBdecl}{\relax}
\BIBdecl

\bibitem{shi_evis_2026}
L.~Shi, R.~Zhang, and Z.~Wang, ``{EVIS}: A physics-grounded event camera plugin
  for {NVIDIA} isaac sim,'' arXiv:2607.08098.

\bibitem{cadena_past_2016}
C.~Cadena, L.~Carlone, H.~Carrillo, Y.~Latif, D.~Scaramuzza, J.~Neira, I.~Reid,
  and J.~J. Leonard, ``Past, {Present}, and {Future} of {Simultaneous}
  {Localization} and {Mapping}: {Toward} the {Robust}-{Perception} {Age},''
  \emph{IEEE Transactions on Robotics}, vol.~32, no.~6, pp. 1309--1332, Dec.
  2016.

\bibitem{ou_efficient_2026}
D.~Ou, J.~J. Hagenaars, M.~R. Jankowski, M.~V.~M. Firlefyn, C.~De~Wagter, F.~T.
  Muijres, J.~Degen, and G.~C. H.~E. de~Croon,
  ``\BIBforeignlanguage{en}{Efficient robot navigation inspired by honeybee
  learning flights},'' \emph{\BIBforeignlanguage{en}{Nature}}, vol. 653, no.
  8116, pp. 1039--1046, May 2026.

\bibitem{gallego_event-based_2022}
G.~Gallego, T.~Delbrück, G.~Orchard, C.~Bartolozzi, B.~Taba, A.~Censi,
  S.~Leutenegger, A.~J. Davison, J.~Conradt, K.~Daniilidis, and D.~Scaramuzza,
  ``Event-{Based} {Vision}: {A} {Survey},'' \emph{IEEE Trans. Pattern Anal.
  Mach. Intell.}, vol.~44, no.~1, pp. 154--180, Jan. 2022.

\bibitem{dinaux_faith_2021}
R.~Dinaux, N.~Wessendorp, J.~Dupeyroux, and G.~C. H. E.~d. Croon, ``{FAITH}:
  {Fast} {Iterative} {Half}-{Plane} {Focus} of {Expansion} {Estimation} {Using}
  {Optic} {Flow},'' \emph{IEEE Robot. Autom. Lett.}, vol.~6, no.~4, pp.
  7627--7634, Oct. 2021.

\bibitem{bhattacharya_monocular_2024}
A.~Bhattacharya, M.~Cannici, N.~Rao, Y.~Tao, V.~Kumar, N.~Matni, and
  D.~Scaramuzza, ``Monocular {Event}-{Based} {Vision} for {Obstacle}
  {Avoidance} with a {Quadrotor},'' Nov. 2024, arXiv:2411.03303.

\bibitem{Rebecq18corl}
H.~Rebecq, D.~Gehrig, and D.~Scaramuzza, ``{ESIM}: an open event camera
  simulator,'' \emph{Conf. on Robotics Learning (CoRL)}, Oct. 2018.

\bibitem{hu_v2e_2021}
Y.~Hu, S.-C. Liu, and T.~Delbruck, ``v2e: From video frames to realistic {DVS}
  events,'' in \emph{2021 {IEEE}/{CVF} Conference on Computer Vision and
  Pattern Recognition Workshops ({CVPRW})}, 2021, pp. 1312--1321.

\bibitem{falanga_dynamic_2020}
D.~Falanga, K.~Kleber, and D.~Scaramuzza, ``Dynamic obstacle avoidance for
  quadrotors with event cameras,'' \emph{Science Robotics}, vol.~5, no.~40, p.
  eaaz9712, 2020.

\bibitem{sanket2020evdodgenet}
N.~J. Sanket, C.~M. Parameshwara, C.~D. Singh \emph{et~al.}, ``{EVDodgeNet}:
  Deep dynamic obstacle dodging with event cameras,'' in \emph{IEEE Int. Conf.
  Robot. Autom. (ICRA)}, 2020.

\bibitem{hagenaars_-device_2025}
J.~J. Hagenaars, Y.~Wu, F.~Paredes-Vallés, S.~Stroobants, and G.~C. de~Croon,
  ``On-{Device} {Self}-{Supervised} {Learning} of {Low}-{Latency} {Monocular}
  {Depth} from {Only} {Events},'' in \emph{2025 {IEEE}/{CVF} {Conference} on
  {Computer} {Vision} and {Pattern} {Recognition} ({CVPR})}, Jun. 2025, pp.
  17\,114--17\,123.

\bibitem{messikommer_approximate_2026}
N.~Messikommer, J.~Xing, L.~Bauersfeld, M.~Cannici, E.~Aljalbout, and
  D.~Scaramuzza, ``Approximate imitation learning for event-based quadrotor
  flight in cluttered environments,'' arXiv:2603.07578.

\bibitem{paredes-valles_fully_2024}
F.~Paredes-Vallés, J.~J. Hagenaars, J.~Dupeyroux, S.~Stroobants, Y.~Xu, and
  G.~C. H.~E. De~Croon, ``Fully neuromorphic vision and control for autonomous
  drone flight,'' \emph{Science Robotics}, vol.~9, no.~90, p. eadi0591, 2024.

\bibitem{zhu_multi_2018}
A.~Z. Zhu, D.~Thakur, T.~Ozaslan, B.~Pfrommer, V.~Kumar, and K.~Daniilidis,
  ``The multi vehicle stereo event camera dataset,'' \emph{IEEE Robot. Autom.
  Lett.}, vol.~3, no.~3, pp. 2032--2039, 2018.

\bibitem{gehrig_dsec_2021}
M.~Gehrig, W.~Aarents, D.~Gehrig, and D.~Scaramuzza, ``{DSEC}: A stereo event
  camera dataset for driving scenarios,'' arXiv:2103.06011.

\bibitem{zhong_event-based_2026}
S.~Zhong, J.~Niu, G.~Gallego, K.~Sun, Y.~Yi, Z.~Miao, D.~Hu, Y.~Wang,
  D.~Scaramuzza, and Y.~Zhou, ``Event-based {SLAM} benchmark for high-speed
  maneuvers,'' arXiv:2604.24033.

\bibitem{gehrig_video_2020}
D.~Gehrig, M.~Gehrig, J.~Hidalgo-Carrió, and D.~Scaramuzza, ``Video to events:
  Recycling video datasets for event cameras,'' in \emph{2020 {IEEE}/{CVF}
  Conference on Computer Vision and Pattern Recognition ({CVPR})}, 2020, pp.
  3583--3592.

\bibitem{zhang_v2ce_2024}
Z.~Zhang, S.~Cui, K.~Chai, H.~Yu, S.~Dasgupta, U.~Mahbub, and T.~Rahman,
  ``V2ce: Video to continuous events simulator,'' in \emph{2024 {IEEE}
  International Conference on Robotics and Automation ({ICRA})}, 2024, pp.
  12\,455--12\,461.

\bibitem{kaiser_towards_2016}
J.~Kaiser, J.~C. Vasquez~Tieck, C.~Hubschneider, P.~Wolf, M.~Weber, M.~Hoff,
  A.~Friedrich, K.~Wojtasik, A.~Roennau, R.~Kohlhaas, R.~Dillmann, and J.~M.
  Zöllner, ``Towards a framework for end-to-end control of a simulated vehicle
  with spiking neural networks,'' in \emph{2016 {IEEE} International Conference
  on Simulation, Modeling, and Programming for Autonomous Robots ({SIMPAR})},
  2016, pp. 127--134.

\bibitem{joubert_iebcs_2021}
D.~Joubert, A.~Marcireau, N.~Ralph, A.~Jolley, A.~van Schaik, and G.~Cohen,
  ``Event camera simulator improvements via characterized parameters,''
  \emph{Frontiers in Neuroscience}, vol.~15, 2021.

\bibitem{pedersen2024gerd}
J.~E. Pedersen, D.~Korakovounis, and J.~Conradt, ``Gerd: Geometric event
  response data generation,'' 2024, arXiv:2412.03259.

\bibitem{de_wagter_autonomous_2014}
C.~De~Wagter, S.~Tijmons, B.~D.~W. Remes, and G.~C. H.~E. de~Croon,
  ``Autonomous flight of a 20-gram {Flapping} {Wing} {MAV} with a 4-gram
  onboard stereo vision system,'' in \emph{2014 {IEEE} {International}
  {Conference} on {Robotics} and {Automation} ({ICRA})}, May 2014, pp.
  4982--4987.

\bibitem{duisterhof_sniffy_2021}
B.~P. Duisterhof, S.~Li, J.~Burgués, V.~J. Reddi, and G.~C. H.~E. de~Croon,
  ``Sniffy {Bug}: {A} {Fully} {Autonomous} {Swarm} of {Gas}-{Seeking} {Nano}
  {Quadcopters} in {Cluttered} {Environments},'' in \emph{2021 {IEEE}/{RSJ}
  {International} {Conference} on {Intelligent} {Robots} and {Systems}
  ({IROS})}, Sep. 2021, pp. 9099--9106.

\bibitem{palossi_64-mw_2019}
D.~Palossi, A.~Loquercio, F.~Conti, E.~Flamand, D.~Scaramuzza, and L.~Benini,
  ``A 64-{mW} {DNN}-{Based} {Visual} {Navigation} {Engine} for {Autonomous}
  {Nano}-{Drones},'' \emph{IEEE Internet Things J.}, vol.~6, no.~5, pp.
  8357--8371, Oct. 2019.

\bibitem{lamberti_sim--real_2024}
L.~Lamberti, E.~Cereda, G.~Abbate, L.~Bellone, V.~J.~K. Morinigo, M.~Barciś,
  A.~Barciś, A.~Giusti, F.~Conti, and D.~Palossi, ``A {Sim}-to-{Real} {Deep}
  {Learning}-{Based} {Framework} for {Autonomous} {Nano}-{Drone} {Racing},''
  \emph{IEEE Robot. Autom. Lett.}, vol.~9, no.~2, pp. 1899--1906, Feb. 2024.

\bibitem{palossi_fully_2022}
D.~Palossi, N.~Zimmerman, A.~Burrello, F.~Conti, H.~Müller, L.~M. Gambardella,
  L.~Benini, A.~Giusti, and J.~Guzzi, ``Fully {Onboard} {AI}-{Powered}
  {Human}-{Drone} {Pose} {Estimation} on {Ultralow}-{Power} {Autonomous}
  {Flying} {Nano}-{UAVs},'' \emph{IEEE Internet Things J.}, vol.~9, no.~3, pp.
  1913--1929, Feb. 2022.

\bibitem{bouwmeester_nanoflownet_2022}
R.~J. Bouwmeester, F.~Paredes-Vallés, and G.~C. H. E.~d. Croon,
  ``{NanoFlowNet}: {Real}-time {Dense} {Optical} {Flow} on a {Nano}
  {Quadcopter},'' Sep. 2022, arXiv:2209.06918.

\bibitem{loquercio_learning_2021}
A.~Loquercio, E.~Kaufmann, R.~Ranftl, M.~Müller, V.~Koltun, and D.~Scaramuzza,
  ``Learning high-speed flight in the wild,'' \emph{Science Robotics}, vol.~6,
  no.~59, p. eabg5810, Oct. 2021.

\bibitem{song_learning_2023}
Y.~Song, K.~Shi, R.~Penicka, and D.~Scaramuzza, ``Learning {Perception}-{Aware}
  {Agile} {Flight} in {Cluttered} {Environments},'' in \emph{2023 {IEEE}
  {International} {Conference} on {Robotics} and {Automation} ({ICRA})}, May
  2023, pp. 1989--1995.

\bibitem{kulkarni_semantically-enhanced_2023}
M.~Kulkarni, H.~Nguyen, and K.~Alexis, ``Semantically-{Enhanced} {Deep}
  {Collision} {Prediction} for {Autonomous} {Navigation} {Using} {Aerial}
  {Robots},'' in \emph{2023 {IEEE}/{RSJ} {International} {Conference} on
  {Intelligent} {Robots} and {Systems} ({IROS})}, Oct. 2023, pp. 3056--3063.

\bibitem{yu_mavrl_2024}
H.~Yu, C.~D. Wagter, and G.~C. H. E.~d. Croon, ``{MAVRL}: {Learn} to {Fly} in
  {Cluttered} {Environments} with {Varying} {Speed},'' Feb. 2024,
  arXiv:2402.08381.

\bibitem{verraest_skydreamer_2025}
A.~Verraest, S.~Bahnam, R.~Ferede, G.~d. Croon, and C.~D. Wagter,
  ``{SkyDreamer}: {Interpretable} {End}-to-{End} {Vision}-{Based} {Drone}
  {Racing} with {Model}-{Based} {Reinforcement} {Learning},'' Oct. 2025,
  arXiv:2510.14783.

\bibitem{yu_avoidbench_2023}
H.~Yu, C.~De~Wagter, and G.~C. H.~E. de~Croon, ``{AvoidBench}: A high-fidelity
  vision-based obstacle avoidance benchmarking suite for multi-rotors,'' in
  \emph{IEEE Int. Conf. Robot. Autom. (ICRA)}, 2023.

\bibitem{serres_optic_2017}
J.~R. Serres and F.~Ruffier, ``Optic flow-based collision-free strategies: From
  insects to robots,'' \emph{Arthropod Structure \& Development}, vol.~46,
  no.~5, pp. 703--717, 2017.

\bibitem{milde_bioinspired_2015}
M.~B. Milde, O.~J.~N. Bertrand, R.~Benosman, M.~Egelhaaf, and E.~Chicca,
  ``Bioinspired event-driven collision avoidance algorithm based on optic
  flow,'' in \emph{IEEE EBCCSP}, 2015.

\end{thebibliography}

\end{document}